\documentclass[letterpaper, 10 pt, conference]{ieeeconf}  % Comment this line out if you need a4paper

\IEEEoverridecommandlockouts                              % This command is only needed if 
\usepackage{amsmath, amssymb}
\let\labelindent\relax
\usepackage{enumerate,enumitem}
\usepackage{graphicx}
\usepackage{algorithm, algorithmic}
\usepackage{cite}
\usepackage{balance}

\newtheorem{ass}{\textbf{Assumption}}

\newtheorem{rem}{\textbf{Remark}}

\newtheorem{prb}{\textbf{Problem}}

\usepackage{lipsum}

\title{\LARGE \bf
Risk-Aware Motion Planning and Control under Unknown Dynamics with Hybrid Observations
}

\author{Zhiquan Zhang$^{\dagger}$ and Melkior Ornik$^{\dagger}$
\thanks{This research was supported by the Air Force Office of Scientific Research under grant number FA9550-23-1-0131.}
\thanks{$^{\dagger}$Zhiquan Zhang and Melkior Ornik are with the University of Illinois Urbana-Champaign, IL, 61801 USA. Emails:
        {\tt\small \{zz121, mornik\}@illinois.edu}}
        }
\begin{document}

\maketitle
\thispagestyle{empty}
\pagestyle{empty}

%%%%%%%%%%%%%%%%%%%%%%%%%%%%%%%%%%%%%%%%%%%%%%%%%%%%%%%%%%%%%%%%%%%%%%%%%%%%%%%%
\begin{abstract}

We consider robotic motion planning and control under unknown dynamics with hybrid state observations, where state measurements are available only in parts of the state space. Existing work combines system identification, predicted reachability, graph search and controller synthesis in a hierarchical framework using local affine approximated models over polytopic state space partitioning, but requires state observations for identification and feedback control. Based on this framework, we address blind regions by selecting nominal dynamics and precomputing open-loop control sequences before observation is lost. Since the true dynamics may differ from the selected nominal model, the robot may exit a blind polytope through an unintended facet. We quantify this transition risk and incorporate the possible outcomes into a stochastic transition system. The high-level planning problem is formulated as a stochastic shortest path problem, whose policy guides controller synthesis. A case study demonstrates that the method guides the robot from an initial state to a target while balancing route efficiency and the risks associated with traversing blind regions.

\end{abstract}

%%%%%%%%%%%%%%%%%%%%%%%%%%%%%%%%%%%%%%%%%%%%%%%%%%%%%%%%%%%%%%%%%%%%%%%%%%%%%%%%
\section{INTRODUCTION}
Robots operating in unfamiliar environments often face uncertainty in their dynamics, arising from environmental interactions or incomplete knowledge of their mechanical parameters~\cite{doi:10.1126/scirobotics.abm6597}. Such uncertainty poses challenges for motion planning: prescribing a sequence of waypoints does not guarantee whether the corresponding motions are achievable under the available control inputs~\cite{9764380}. System identification provides information about the unknown dynamics that can support reachability analysis and controller synthesis. However, the choice of control inputs determines not only the robot's motion but also the information collected about its dynamics~\cite{8759089}. Motion planning must therefore balance exploration for system dynamics with exploitation of the information already available. In particular, visiting more informative regions may require additional travel time, whereas a more direct route toward the target may provide fewer opportunities to reduce model uncertainty~\cite{8759089, doi:10.1177/0278364914533443}.

Recent works~\cite{11245807,author2026paper} addressed these challenges through a hierarchical motion planning and control framework for systems with unknown dynamics. The robot's state space is partitioned into convex polytopes, and the nonlinear dynamics within each polytope are locally approximated by an affine model. Facet reachability analysis is then used to construct a directed graph whose nodes represent polytopes and edges describe transitions between adjacent polytopes. The framework combines affine system identification, reachability prediction, graph construction, and motion planning in an online procedure. As new information about the dynamics becomes available, the local models and graph are updated to guide further planning and control. This procedure balances exploration (system identification) and exploitation (motion planning and control), enabling the robot to find a feasible path towards the target.

However, the framework described above relies on state observations for online system identification and feedback control. In this paper, we consider robotic motion planning and control problems under unknown system dynamics with hybrid state observations, where exact state measurements are available in some regions and unavailable in others.

In practice, visual occlusions, limited fields of view and poor camera visibility, communication interruptions, etc., may restrict state measurements~\cite{8593739, 7070677}. Existing model-based approaches account for limited observations by predicting state evolution and incorporating the resulting uncertainty into motion planning~\cite{doi:10.1177/0278364912456319, 8479330, 8543823}. Related control methods use state prediction to maintain bounded tracking errors while the system moves between regions with and without feedback~\cite{8671728}. A common requirement of these approaches is the availability of the motion model for predicting the unobserved state. Learning-based methods have also been developed for unknown systems under feedback interruptions and intermittent observations~\cite{9903320, ZHU20232265}. These approaches use previously collected input-output data for prediction and controller design. However, their online control decisions do not explicitly balance the acquisition of additional dynamics information against control performance. This tradeoff is important in robotic navigation, where visiting a more informative region may improve subsequent planning but require additional traversal time.

 In this work, we extend the motion planning and control framework in~\cite{11245807,author2026paper} to this setting. The absence of state measurements in blind regions prevents direct local system identification and state feedback control. A nominal dynamics model can be selected from the admissible parameter set and used to synthesise an open-loop control sequence before the robot enters a blind regions. This sequence can then be executed without state measurements. However, discrepancies between the nominal model and the true dynamics may cause the robot to leave through an unintended facet and enter an unintended tile. Although excluding these regions avoids the risks associated with unintended tile transitions, it may also rule out more efficient routes. We therefore introduce a quantitative measure of unintended tile transition risk and incorporate the possible transition outcomes into a stochastic transition system. The planning approach accounts for both traversal efficiency and the consequences of unintended transitions.

The remainder of this paper is organized as follows. Section~\ref{sec:preliminary} introduces necessary preliminaries and notations used throughout the paper. Section~\ref{sec:problemFormulation} formally states the problem considered in this work. Section~\ref{sec:review} briefly reviews the motion planning and control frameworks proposed in~\cite{11245807,author2026paper}, which serve as the basis for the proposed method. Section~\ref{sec:hybridObservation} presents the proposed approach for systems with hybrid state observations. Section~\ref{sec:caseStudy} provides a case study to demonstrate the proposed method. Finally, section~\ref{sec:conclusion} concludes the paper.

\section{PRELIMINARIES AND NOTATIONS}\label{sec:preliminary}
In this section, we present the preliminaries and notation required for the subsequent sections. Specifically, we introduce stochastic transition systems and stochastic shortest path planning problems, and formally define some notations and terminology used throughout the paper.

\subsection{Stochastic Transition System}\label{subsec:stochasticTransition}
A \textit{stochastic transition system} is defined as a tuple $\mathcal{T}=(\mathcal{S}, \mathcal{A}, P, c, s_0)$, where $\mathcal{S}$ is the finite set of \textit{states}, $\mathcal{A}$ is the finite set of \textit{actions}, and $s_0 \in \mathcal{S}$ is the initial state. The \textit{transition probability function} $P$ is defined as $P:\mathcal{S}\times \mathcal{A} \times \mathcal{S} \rightarrow [0, 1]$. For a state $s \in \mathcal{S}$ and an admissible action associated with this state $a \in \mathcal{A}(s)$, $P(s'|s, a)$ represents the probability of transition from $s$ to $s'$ after applying action $a$. Naturally, $P(s'|s, a)$ satisfies $\sum_{s'\in \mathcal{S}} P(s'|s, a) = 1$. The \textit{transition cost function} $c$ is defined as $c:\mathcal{S}\times\mathcal{A}\times\mathcal{S}\rightarrow \mathbb
R_{\ge 0}$, and $c(s', a, s)$ denotes the cost when the system transitions from $s$ to $s'$ under action $a$. A transition is \textit{deterministic} if for specific $s$, $s'$ and $a$ such that $P(s'|s, a) = 1$, and is \textit{stochastic} otherwise. Under this definition, the stochastic transition system considered in this paper can be equivalently viewed as a \textit{Markov decision process} (MDP).

A (stationary, deterministic) \textit{policy} specifies how an action is selected at each state, which is defined as a mapping $\pi:\mathcal{S}\rightarrow\mathcal{A}$. Given a policy $\pi$, a run of the stochastic transition system is a sequence $r = s_0a_0s_1a_1s_2\ldots$, where $a_k = \pi(s_k)$ and $s_{k+1}$ is sampled by $P(s_{k+1}|s_k, a_k)$. The cumulative cost is defined as the sum of the costs along all the transitions $C(r) = \sum_{k}c(s_k, a_k, s_{k+1})$. As introduced below, the problems considered in this paper contain an absorbing terminal state, and we consider proper policies under which this terminal state is reached almost surely. Therefore, the cumulative cost is accumulated only until the terminal state is reached.

\subsection{Stochastic Shortest Path Planning}\label{subsec:SSPP}
A \textit{stochastic shortest path (SSP) problem} is defined on a stochastic transition system with a target state $s_g\in \mathcal{S}$. The target state is assumed to be absorbing and yields no further cost after it is reached, i.e., $P(s_g|s_g, a) = 1$, $c(s_g, a, s_g) = 0$. Starting from an initial state $s_0$, the objective of the SSP problem is to find a policy $\pi^*$ that reaches $s_g$ while minimizing the expected cumulative cost. Let $\tau_g = \inf\{k\ge0:s_k = s_g\}$, which denotes the first hitting time of the target state. The expected cumulative cost under a policy $\pi$ is 
\begin{equation}\label{eq:Jpis0}
    J^\pi(s_0) = \mathbb{E}_\pi\left[\sum_{t=0}^{\tau_g-1}c(s_t, a_t, s_{t+1}) \Big|s_0\right].
\end{equation}
The optimal value function associated with each state is then defined as
\begin{equation}
    J^*(s) = \min_\pi J^\pi(s).
\end{equation}

A policy $\pi$ is called proper if it guarantees that the target set is reached almost surely, i.e., $P_\pi(\tau_g < \infty|s_0) = 1$. The optimal value function satisfies the Bellman optimality equation
\begin{equation}
    J^*(s) = 
        \min_{a\in \mathcal{A}(s)} \sum_{s'\in \mathcal{S}}P(s'|s, a)[c(s, a, s') + J^*(s')], s \neq s_g,
\end{equation}
with $J^*(s) = 0$ when $s = s_g$.

To make the problem considered later is well-defined, we make two standard assumptions: (i) there exists at least one proper policy; (ii) every improper policy has infinite expected cumulative cost for at least one initial state.

\subsection{Facet Reachability and Reach Control}\label{sec:facetReachability}
The stochastic transition system and SSP provide the high-level planning model, while facet reachability and reach control determine and realize transitions between neighboring polytopic regions. In this subsection, we briefly introduce the facet reachability conditions and the corresponding reach controller used for the motion within a polytopic tile. These conditions determine whether a specified facet can be reached in finite time. These results serve as theoretical foundations for the methods developed later in this paper.

Consider a bounded and convex \textit{polytope} $P \subset \mathbb{R}^n$ with a set of \textit{vertices} $\{v_j\}_{j \in \mathcal{V}}$ and \textit{facets} $\{F_i\}_{i \in \mathcal{F}}$. The polytope is defined by $P = \{x\in \mathbb{R}^n|H_px \leq h_p\}$, where $H_p \in \mathbb{R}^{N_f\times n}$ and $h_p\in \mathbb{R}^{N_f}$. Let $n_i$ denote the outward normal vector of facet $F_i$, and let $W_j\subseteq \mathcal{F}$ denote the set of facets that contain vertex $v_j$. Consider an affine system $\dot x = Ax + Bu + c$ in the polytope, where $u \in \mathcal{U}$. 

We firstly define the \textit{Reach Control Problem} (RCP) considered in this paper. Let $F_e$ be a prescribed target facet with outward normal $n_e$. Following~\cite{broucke2014reach, habets2004control}, $F_e$ is said to be \textit{reachable} under affine state-feedback control if, for every initial state $x_0 \in P$, there exist an admissible affine controller $u(x) = Fx + g$ and a finite time $T_0>0$, such that $x(t) \in P,\ \forall t\in[0, T_0)$, $x(T_0) \in F_e$, and $n_e^\top \dot x(T_0)>0$, while $u(x(t)) \in \mathcal{U}$ for all $t \in [0, T_0]$.

We next introduce sufficient conditions for establishing this facet reachability. Let $\mathcal{V}\subset\mathcal{V}$ denote the set of vertices lying on $F_e$. If there exist admissible control inputs on vertices of $P$ satisfying the following conditions, a reach controller towards $F_e$ can be constructed. For each vertex $v_j$ on the target facet, $j \in \mathcal{V}_e$, 
\begin{equation}\label{eq:reachOriginal1}
    n_e^\top(Av_j + Bu_j + c)>0,
\end{equation}
and 
\begin{equation}\label{eq:reachOriginal2}
    n_i^\top (Av_j + Bu_j + c) \leq 0,\ \forall i \in W_j\backslash\{e\}.
\end{equation}
For each vertex not on the target facet, $j \in \mathcal{V}\backslash \mathcal{V}_e$, 
\begin{equation}\label{eq:reachOriginal3}
    n_i^\top (Av_j + Bu_j + c) \leq 0,\ \forall i \in W_j,
\end{equation}
and 
\begin{equation}\label{eq:reachOriginal4}
    n_e^\top (Av_j + Bu_j + c)>0. 
\end{equation}

Once a feasible set of vertex control inputs satisfying the above reachability conditions is obtained, we can construct the corresponding reach controller to reach $F_e$. For a simplex with vertices $v_i, \ldots,v_{n+1}$, let the corresponding feasible vertex controls be $\bar u_1, \ldots, \bar u_{n+1}$. The affine reach controller is written as
\begin{equation}\label{eq:controller}
    u(x) = F_lx + g_l,
\end{equation}
where $F_l$ and $g_l$ are determined from
\begin{equation}\label{eq:controllerParameter}
    \begin{bmatrix}
F_l \mid g_l
\end{bmatrix}
\begin{bmatrix}
\bar v_{l1} & \cdots & \bar v_{l(n+1)} \\
1 & \cdots & 1
\end{bmatrix}
=
\begin{bmatrix}
\bar u_{l1} & \cdots & \bar u_{l(n+1)}
\end{bmatrix}.
\end{equation}

For a general polytope, the region can be triangulated into simplices, and an affine controller is constructed on each simplex following the same procedure.

\subsection{Notations}
Throughout the paper, $\|\cdot\|$ denotes 2-norm for vectors and Frobenius norm for matrices.

\section{PROBLEM FORMULATION}\label{sec:problemFormulation}
Consider an agent operating under continuous-time nonlinear control affine dynamics
\begin{equation}\label{eq:dynamics}
    \dot x = f(x) +g(x)u,
\end{equation}
where $x \in \mathbb{R}^n$ is the system state and $u \in \mathbb{R}^m$ is the control input. The functions $f:\mathbb{R}^n \rightarrow \mathbb{R}^n$ and $g:\mathbb{R}^n \rightarrow \mathbb{R}^{n \times m}$ represent the drift vector field and input matrix, respectively. We make the following assumptions to formulate the problem.
\begin{ass}\label{ass:constaints}
    \textbf{(Constrained state space and control inputs)} We assume that the system state is constrained to a polytope $P_s \subset \mathbb{R}^n$, and the control input is restricted to a polytope $P_u \subset \mathbb{R}^m$.
\end{ass}
\begin{ass}\label{ass:Lipschitz}
    \textbf{(Lipschitz continuity)} We assume that the functions $f(x)$ and $g(x)$ are a priori unknown except for their dimensions. However, assuming $f(x)$ is differentiable with respect to $x$, $\nabla_x f(x)$ and $g(x)$ are Lipschitz continuous on $P_s$, and there exists known constants $\mathcal{L}_{df}$ and $\mathcal{L}_g$ such that $\|\nabla_x f(x_1) - \nabla_x f(x_2)\| \leq \mathcal{L}_{df}\|x_1 - x_2\|$ and $\|g(x_1) - g(x_2)\| \leq \mathcal{L}_g\|x_1 - x_2\|$ for any $x_1,x_2 \in P_s$.
\end{ass}
% \begin{ass}\label{ass:hybridObservation}
%     \textbf{(Hybrid observation)} The state space $P_s$ contains both observed and unobserved regions. Let $S_o \subseteq P_s$ denote the subset in which perfect state measurements are available, and let $S_b = P_s\backslash P_o$ denote the subset in which no state measurement is available. The partition of $P_s$ into $S_o$ and $S_b$ is assumed to be a priori known to the agent.
% \end{ass}

Assumption~\ref{ass:constaints} captures physical limitations on the state and control input, such as workspace boundaries and actuator saturation. Assuption~\ref{ass:Lipschitz} is satisfied by a broad class of smooth nonlinear systems. This assumption guarantees that the system dynamics are locally well-posed and admit unique trajectories~\cite{9764380}. 

The objective of this paper is formulated as follows:
\begin{prb}\label{prb1}
    Consider an agent governed by the unknown nonlinear control affine system~\eqref{eq:dynamics}, satisfying the stated Lipschitz continuity assumptions. The state space $P_s$ is partitioned into an observed region $S_o \subseteq P_s$, where perfect state measurements are available, and a blind region $S_b  = P_s\backslash S_o$, where no state measurements are available. The partition of $P_s$ into $S_o$ and $S_b$ is assumed to be a priori known to the agent. Given an initial state $x_0\in P_s$ and a target state $x^*\in P_s$, design an admissible control strategy $u$ such that the resulting trajectory reaches $x^*$ in finite time.
\end{prb}
We do not assume that $P_s$ is positively invariant under arbitrary control policies; rather, only control strategies that keep the resulting trajectory inside $P_s$ are considered admissible.
\begin{ass}\label{ass:initialTarget}
    We assume that the initial state and target state both lie within the observed region, i.e., $x_0,x^* \in P_o$.
\end{ass}

Assumption~\ref{ass:initialTarget} ensures that the system state is exactly known at the beginning of the task and can be directly verified when reaching the target. This assumption avoids ambiguity in initializing and determining successful task completion.

\section{REVIEW OF MOTION PLANNING AND CONTROL UNDER UNKNOWN DYNAMICS}\label{sec:review}

In this section, we briefly review the motion planning and control framework under unknown dynamics via predicted reachability introduced in~\cite{11245807,author2026paper}. Previous work considers the fully observed special case of Problem~\ref{prb1}, where state measurements are available throughout the state space. Even this simpler setting remains challenging because the system dynamics are unknown, and system properties such as reachability cannot be determined. The agent therefore needs to acquire information about the dynamics while planning and executing motions towards the target. The complete formulation and technical details can be found in~\cite{11245807,author2026paper}.

The work of~\cite{11245807,author2026paper} partitions the state space $P_s$ into a finite collection of convex polytopes $\{P_l\}_{l \in L}$ and locally approximates the nonlinear dynamics within each polytope by an affine model.
\begin{equation}\label{eq:approximateDynamics}
    \dot x \approx \bar A_l x + \bar B_l u + \bar c_l.
\end{equation}
The polytopic partition is then represented by a directed graph $\mathcal{G}_s = (V_s, E_s, w_s)$, where each node corresponds to a polytope and a directed edge $(l, l')$ represents a feasible transition from $P_l$ to an adjacent polytope $P_{l'}$ through a reachable facet. The transition is determined by facet reachability. Since the system dynamics are initially unknown, these reachability relations and graph edges are not known. Based on this representation, Problem~\ref{prb1} is reformulated as determining feasible transition sequences and constructing the corresponding reach controllers under initially unknown local affine dynamics. The agent is guided from $x_0\in P_0$ to the target polytope $P_*$ in finite polytope, then a local controller drives the state to $x^* \in P_*$.

\subsection{System Identification and Predicted Reachability}\label{subsec:systemId}
When the agent enters a polytope where no model information has been acquired, small excitation inputs and the corresponding state responses are used in least-squares regression to estimate the local affine model parameters $(\bar A_l, \bar B_l,\bar c_l)$~\cite{ornik2019control}. Facet reachability is then examined using conditions in Section~\ref{sec:facetReachability}. This analysis is extended to unidentified polytopes using the available dynamics information~\cite{11245807}.

Consider an identified polytope $P_l$ with affine parameters $(\bar A_l, \bar B_l, \bar c_l)$, and another polytope $P_{l'}$ whose local affine parameters $(\bar A_{l'}, \bar B_{l'}, \bar c_{l'})$ remain unknown. Under the Lipschitz continuity assumption, their differences satisfy
\begin{equation}\label{eq:dynamicsBounds}
    \|\bar A_{l'} - \bar A_l\| \leq \varepsilon_A,\ \|\bar B_{l'} - \bar B_l\| \leq \varepsilon_B,\ \|\bar c_{l'} - \bar c_l\| \leq \varepsilon_c,
\end{equation}
where, for system identification points $x_l$ and $x_{l'}$, $\varepsilon_A = \mathcal{L}_{df}\|x_{l'} - x_l\|$, $\varepsilon_B = \mathcal{L}_{g}\|x_{l'} - x_l\|$, and $\varepsilon_c = \frac{1}{2}\mathcal{L}_{df}\|x_{l'} - x_l\|^2 + \mathcal{L}_{df}\|x_{l'} - x_l\|\|x_{l'}\|$.

Let $\mathcal{I}_+$ and $\mathcal{I}_-$ denote the facet index sets corresponding to strict positive inequalities in reachability conditions. If local dynamics of $P_{l'}$ were known, the conditions would be written as
\begin{equation}
    \begin{aligned}
        &n_{l'i}^\top (\bar A_{l'}v_{l'j} + \bar B_{l'} u_{l'j} + \bar c_{l'}) >0,\ i \in \mathcal{I}_+,\\
        &n_{l'i}^\top (\bar A_{l'}v_{l'j} + \bar B_{l'} u_{l'j} + \bar c_{l'}) \leq 0,\ i \in \mathcal{I}_-,
    \end{aligned}
\end{equation}
with $u_{l'j}\in P_u$. For unknown dynamics, the bounds~\eqref{eq:dynamicsBounds} are used to construct conservative reachability conditions. For a given sign configuration of $u_{l'j}$, define
\begin{equation}
\scalebox{0.99}{$    \begin{aligned}
        \bar B^+_{l'} &= n_{l'i}^\top \bar B_l+ \left[\textbf{sgn}(u_{l'j}^1)\varepsilon_B\|n_{l'i}\|\ \cdots \textbf{sgn}(u_{l'j}^m)\varepsilon_B\|n_{l'i}\|\right ],\\
        \bar B^-_{l'} &= n_{l'i}^\top \bar B_l-\left[\textbf{sgn}(u_{l'j}^1)\varepsilon_B\|n_{l'i}\|\ \cdots\ \textbf{sgn}(u_{l'j}^m)\varepsilon_B\|n_{l'i}\|\right ].\\
    \end{aligned}$}
\end{equation}
A sufficient condition for the selected facet of $P_{l'}$ to be reachable is the feasibility of
\begin{equation}\label{eq:reachableCondition}
\scalebox{0.94}{$\begin{aligned}
        &\bar B^-_{l'}u_{l'j} + n^\top_{l'i}(\bar A_l v_{l'j} + \bar c_l) > \|n_{l'i}\|(\varepsilon_A\|v_{l'j}\| + \varepsilon_c),\ i \in \mathcal{I}_+,\\
        &\bar B^+_{l'}u_{l'j} + n^\top_{l'i}(\bar A_l v_{l'j} + \bar c_l) \leq -\|n_{l'i}\|(\varepsilon_A\|v_{l'j}\| + \varepsilon_c),\ i \in \mathcal{I}_-,
\end{aligned}$}
\end{equation}
with $u_{l'j} \in P_u$. These inequalities enforce the reachability conditions for every local affine model satisfying the parameter bounds in~\ref{eq:dynamicsBounds}.

Conversely, an expanded set of reachability inequalities is
\begin{equation}\label{eq:nonreachableCondition}
\scalebox{0.94}{$\begin{aligned}
        &\bar B^+_{l'}u_{l'j} + n^\top_{l'i}(\bar A_l v_{l'j} + \bar c_l) > -\|n_{l'i}\|(\varepsilon_A\|v_{l'j}\| + \varepsilon_c),\ i \in \mathcal{I}_+,\\
        &\bar B^-_{l'}u_{l'j} + n^\top_{l'i}(\bar A_l v_{l'j} + \bar c_l) \leq \|n_{l'i}\|(\varepsilon_A\|v_{l'j}\| + \varepsilon_c),\ i \in \mathcal{I}_-.
\end{aligned}$}
\end{equation}
The framework classifies the selected facet as unreachable if this inequality set is infeasible. If neither condition provides a conclusive result, facet reachability remains undetermined.

Since $\bar B^+_{l'}$ and $\bar B^-_{l'}$ depend on the signs of the components of $u_{l'j}$, enumerating all sign configurations of an $m$-dimensional input requires up to $2^m$ affine feasibility problems.

\subsection{Non-uniform State Space Partitioning}\label{subsec:nonuniformPartition}
A uniform partition of an $n$-dimensional state space leads to a rapid growth in the number of polytopes as the partition resolution increases. The work of~\cite{author2026paper} introduced an adaptive non-uniform partition to concentrate finer cells in regions relevant to the current planning task while maintaining coarser cells elsewhere.

Let $x_c$ and $x^*$ denote the state of the agent at the current planning iteration and target state respectively. For simplicity, we assume that the state space $P_s$ is a hypercube and employ cubical partitions. The method starts from a coarse cubical partition and recursively subdivides every cube intersected by the line segment connecting $x_c$ and $x^*$. Each selected cube is divided equally along all coordinate axes into $2^n$ smaller cubes until the intersected cubes reach the minimum side length $h_{\rm min}$. The procedure is repeated as $x_c$ changes. 

\subsection{Graph Construction and Motion Planning Framework}\label{subsec:graphConstruction}
The polytopic partition and the predicted reachability results are used to construct a directed weight graph $\mathcal{G}_s = (V_s, E_s, w_s)$, where each vertex represents a polytope. A directed edge from $P_l$ to an adjacent polytope $P_{l'}$ is included when their shared facet is certified as reachable. If the condition~\eqref{eq:nonreachableCondition} is infeasible, the corresponding transition is excluded. The remaining transitions are retained as uncertain edges. When the exit facet of $P_l$ is larger than the corresponding entry facet of $P_{l'}$, the transition is also treated as uncertain.

For a transition whose reachability is confirmed with exit facet $n_e$, vertices $v_{lj}$, and associated controls $u_{lj}$, the edge weight is determined by the upper bound on the time required to leave the current polytope $T_l \leq (\beta - \alpha)/c_1$, where $\alpha = \min_j\{n_e^\top v_{lj}\}$, $\beta = \max_j \{n_e^\top v_{lj}\}$, and $c_1 = \min_j\{n_e^\top \bar A_l v_{lj} + \bar B_l u_{lj} + \bar c_l\}$. For a polytope whose dynamics have not been identified with certified reachability, $c_1$ is replaced by the worst-case lower bound on the velocity along the target facet normal over all dynamics satisfying~\eqref{eq:dynamicsBounds}.

Uncertain edges are assigned exploratory weights based on the information that may be obtained after traversing them. Let $p_e$ denote the prior probability that an uncertain edge $e$ exists. Its uncertainty is measured by the entropy $H(e) = -[p_e\log p_e + (1-p_e)\log(1-p_e)]$. We assume that when no prior information is available, $p_e = 0.5$. The expected information gain is given by~\cite{author2026paper} as
\begin{equation}
    \mathbb{E}[IG_e] = p_e \sum_{e'\in\mathcal{E}(l')} H(e'),
\end{equation}
where $\mathcal{E}(l')$ denotes the uncertain outgoing edges of the reached polytope $P_{l'}$. The corresponding edge weight is selected as
\begin{equation}\label{eq:uncertainWeights}
    w(e) = \frac{C_u l_u}{1 + \beta_u \mathbb{E}[IG_e]},
\end{equation}
where $l_u$ denotes the length of the corresponding cell along the transition direction, $C_u>0$ is a scaling constant, and $\beta_u>0$ controls the influence of exploration.

At each planning step, the graph is updated using new dynamics information, and a shortest path search is performed from the current polytope to the target polytope. A reach controller executes the first transition on the path. The procedure repeats until the target polytope is reached, after which the agent is driven to the target state.

\section{MOTION PLANNING AND CONTROL WITH HYBRID OBSERVATION}\label{sec:hybridObservation}
In this section, we extend the motion planning and control framework reviewed in the previous section to the hybrid observation setting. The agent may enter regions in which state observations are unavailable. Rather than naively treating these regions as obstacles, we explicitly account for the uncertainty associated with traversing them and incorporate the resulting transitions into the high-level problem.

Consider an unobserved tile $P_l \subset S_b$ and suppose that both the entry state and the local affine dynamics are known exactly. In this case, the absence of state measurements inside the tile does not prevent execution of the reach controller. Assume that the local dynamics are represented by~\eqref{eq:dynamics} and an affine state feedback controller~\eqref{eq:controller} has been synthesized. Starting from the known entry state $x^e$, the closed loop state trajectory and the corresponding control input can be computed before the agent enters the tile as
\begin{equation}\label{eq:openLoopcontroller}
\scalebox{0.9}{$
\begin{aligned}
    &\dot x = (\bar A_l + \bar B_l F_l)x + \bar B_lg_l+c_l,\ x(0) = x^e\\
    \Rightarrow & x = e^{(\bar A_l+\bar B_lF_l)t} x_0 + \int_0^t e^{(\bar A_l+\bar B_l \bar F_l)(t-\tau)}(\bar B_l g_l + \bar c_l)d\tau\\
    \Rightarrow & u = F_l[e^{(\bar A_l+\bar B_lF_l)t} x_0 + \int_0^t e^{(\bar A_l+ \bar B_l F_l)(t-\tau)}(\bar B_l g_l + \bar c_l)d\tau] + g_l,
\end{aligned}
$}
\end{equation}
where $\bar A_l$, $\bar B_l$, $\bar c_l$, $F_l$, $g_l$ are given by~\eqref{eq:approximateDynamics} and~\eqref{eq:controller}. Therefore, if the local model is known perfectly, the feedback controller can be converted into a precomputed open-loop control trajectory and executed without state measurements inside the tile.

The main difficulty is that the local affine dynamics of an unobserved tile cannot be identified directly, and also the true dynamics are not affine. Using the Lipschitz bounds and the affine models identified in observed regions, the unknown parameters can be bounded by~\eqref{eq:dynamicsBounds}. An open-loop control trajectory synthesized from a nominal model may lead the system to a facet different from the intended one. Treating every unobserved tile as an obstacle would avoid this uncertainty but may exclude shorter or more preferable routes. We instead quantify the risk of unintended tile transitions, construct a stochastic graph that includes transitions through unobserved regions, and formulate Problem~\ref{prb1} as a stochastic shortest path problem.

\subsection{Risk of Unintended Tile Transition}\label{subsec:riskUnintended}
In this subsection, we introduce a quantitative metric for characterizing the risk of unintended tile transitions. Let $\Theta$ denote the admissible set of local affine dynamics parameters $(A,B,c)$ associated with a given blind tile. Since the true dynamics are unknown, a nominal model must be selected from $\Theta$ for controller synthesis. For a prescribed target facet, different choices of the nominal model and the corresponding controller may result in different sets of true dynamics for which the intended tile transition can be guaranteed. Therefore, we first select the nominal dynamics and the associated vertex control inputs to maximize the volume of this certified parameter set. Assuming that the true dynamics parameters are uniformly distributed over $\Theta$, a larger certified volume roughly corresponds to a higher probability of successfully exiting through the intended facet.

Let $(A^*, B^*, c^*)$ denote the center of the dynamics uncertainty set. The set of all possible affine dynamics is then represented by
\begin{equation}
\scalebox{0.88}{$
    \Theta = \{(A, B, c):\|A - A^*\|\leq \varepsilon_A, \|B - B^*\|\leq \varepsilon_B, \|c - c^*\|\leq \varepsilon_c\},$}
\end{equation}
where $\varepsilon_A$, $\varepsilon_B$, $\varepsilon_c$ are the uncertainty bounds obtained from~\eqref{eq:dynamicsBounds}. For a prescribed target facet $F_e$, we seek a subset of $\Theta$ such that $F_e$ remains reachable for all dynamics within this subset under a selected set of vertex control inputs. The resulting certified set $\Theta_{\rm reach}$ is given by
\begin{equation}\label{eq:thetaReach}
    \Theta_{\rm reach} = \left\{
(A,B,c)\; \middle|\;
\begin{aligned}
\|A-(A^* + C_A)\| &\leq \rho_A \varepsilon_A,\\
\|B-(B^* + C_B)\| &\leq \rho_B \varepsilon_B,\\
\|c-(c^* + C_c)\| &\leq \rho_c \varepsilon_c
\end{aligned}
\right\}.
\end{equation}
To guarantee $\Theta_{\rm reach} \subseteq \Theta$, we impose
\begin{equation}\label{eq:constraintsRadius}
    \begin{aligned}
        &\|C_A\| + \rho_A\varepsilon_A \leq \varepsilon_A,\ 0\leq \rho_A \leq 1,\\
        &\|C_B\| + \rho_B\varepsilon_B \leq \varepsilon_B,\ 0\leq \rho_B \leq 1,\\
        &\|C_c\| + \rho_c\varepsilon_c \leq \varepsilon_c,\ 0\leq \rho_c \leq 1.\\
    \end{aligned}
\end{equation}

Let $d_A = n^2$, $d_B = nm$, $d_c = n$ denote the dimensions of the parameter spaces of $A$, $B$ and $c$. Since the volume of $\Theta_{\rm reach}$ is proportional to $(\rho_A \varepsilon_A)^{d_A}(\rho_B \varepsilon_B)^{d_B}(\rho_c \varepsilon_c)^{d_c}$~\cite{schneider2014convex}, maximizing its volume is equivalent to maximizing $d_A\log \rho_A + d_B\log \rho_B + d_c\log \rho_c$

\begin{equation}\label{eq:bigOptimization}
\scalebox{0.88}{$\begin{aligned}
\max_{C_*, \rho_*, u_j}\ & d_A \log \rho_A + d_B \log\rho_B + d_c \log \rho_c\\
\textrm{s.t.}\
& n_i^\top[(A^* + C_A)v_j + (B^* + C_B)u_j + (c_0 + C_c)] + \|n_i\|(\rho_A\\&\ \ \ \ \ \varepsilon_A\|v_j\|_2 + \rho_B\varepsilon_B \|u_j\|_2 + \rho_c\varepsilon_c) \leq 0, \forall j, i \in W_j \backslash\{e\},\\
& n_e^\top[(A^* + C_A)v_j + (B^* + C_B)u_j + (c_0 + C_c)] \\&\ \ \ \ \ - \|n_i\|(\rho_A\varepsilon_A\|v_j\|_2 + \rho_B\varepsilon_B \|u_j\|_2 + \rho_c\varepsilon_c) > 0, \forall j,\\
&\|C_A\| + \rho_A\varepsilon_A \leq \varepsilon_A,\ 0\leq \rho_A \leq 1,\\
&\|C_B\| + \rho_B\varepsilon_B \leq \varepsilon_B,\ 0\leq \rho_B \leq 1,\\
&\|C_c\| + \rho_c\varepsilon_c \leq \varepsilon_c,\ 0\leq \rho_c \leq 1,\\
& u_j \in \mathcal{U}, \forall j.
\end{aligned}$}
\end{equation}
Let $C_A^*$, $C_B^*$, $C_c^*$, $\rho_A^*$, $\rho_B^*$, $\rho_c^*$, $\{u_j^*\}$ denote an optimal solution. The nominal dynamics used for controller synthesis are then selected as
\begin{equation}
    \hat A = A^* + C_A^*,\ \hat B = B^* + C_B^*,\ \hat c = c^* + C_c^*,
\end{equation}
and the corresponding $\Theta_{\rm reach}^*$ by substituting the optimal variables into~\eqref{eq:thetaReach}.

The solution to~\eqref{eq:bigOptimization} also gives the vertex control inputs $u_j^*$, which are used to construct the reach controller associated with the nominal dynamics.

Once the nominal model and the vertex control inputs are fixed, the transition probability associated with each facet can be evaluated. For facet $F_k$, define
\begin{equation}\label{eq:thetaReach}
\scalebox{0.93}{$\Theta_{\rm reach}^k(u) = \left\{
(A,B,c)\in \Theta\; \middle|\;
\begin{aligned}
&n_i^\top (Av_j + Bu_j + c) \leq 0,\\&\ \ \ \ \ \ \ \ \ \ \ \ \ \ \ \  \forall j, i \in W_j \backslash\{k\},\\
&n_k^\top (Av_j + Bu_j + c) > 0,\ \forall j.\\
\end{aligned}
\right\}.$}
\end{equation}
This compact form is equivalent to the facet reachability condition~\eqref{eq:reachOriginal1}-\eqref{eq:reachOriginal4}. For fixed $u_j$, they are constraints affine in $A$, $B$ and $c$.

Since the reachability conditions are sufficient, $\Theta^k_{\rm reach}(u)$ is generally an inner approximation of the set of dynamics that actually lead to $F_k$. The corresponding approximated transition probability to $F_k$ is represented by
\begin{equation}
    p_k = \frac{{\rm Vol}(\Theta^k_{\rm reach}(u))}{{\rm Vol} (\Theta)}.
\end{equation}
The same calculation is applied to the intended facet and each unintended facet to obtain the transition probabilities used in the stochastic graph. Naturally, if $\sum_k p_k<1$, the transition probabilities are normalized so that they sum to one.

\subsection{Stochastic Graph Construction}
We construct the stochastic graph for motion planning using the stochastic transition system that is introduced in Section~\ref{subsec:stochasticTransition}. Particularly, the partitioned state space and the transitions between tiles are represented by $\mathcal{T} = (V, \mathcal{A}, P, c, v_0)$, where $V$ is the finite set of tiles, $\mathcal{A}$ is the set of admissible facet transition actions, and $v_0$ is the tile containing the initial state $x_0$. According to the observation availability, the set $V$ is divided into
\begin{equation}
    V = V_{\rm obs} \cup V_{\rm blind},\ V_{\rm obs} \cap V_{\rm blind} = \varnothing.
\end{equation}
A tile is classified as blind if any portion of it intersects the blind region $S_b$. Accordingly,
\begin{equation}
    V_{\rm blind} = \{v_i \in V|v_i \cap S_b \neq \varnothing\},\ V_{\rm obs} = V\backslash V_{\rm blind}.
\end{equation}
For each tile $v_i \in V$, let $\mathcal{A}(v_i) \subseteq \mathcal{A}$ denote its admissible action set. An action $a \in \mathcal{A}(v_i)$ specifies a target facet of $v_i$, and the adjacent tile corresponding to this facet is denoted by $\tau(v_i, a)$. The admissible actions are determined from the facet reachability and predicted reachability analysis described in Section~\ref{subsec:systemId}.

The partition procedure follows the non-uniform partitioning strategy introduced in~\ref{subsec:nonuniformPartition}, with the additional requirement that every tile intersecting $S_b$ is refined to the minimum side length $h_{\rm min}$. This additional refinement reduces the conservativeness introduced by classifying a partially blind tile as entirely blind and provides finer resolution near the boundary between observed and blind regions.

For an observed tile $v_i \in V_{\rm obs}$, the local dynamics can be identified and facet transitions are deterministic if the corresponding reachability conditions are satisfied. Thus, 
\begin{equation}
    P(\tau(v_i, a)|v_i, a) = 1,\ P(v_j|v_i, a) = 0,\ \forall v_j \neq \tau(v_i, a).
\end{equation}
\begin{rem}
    The uncertain transitions related to the predicted facet reachability become certified once the agent enters these observed tiles. Thus, the transitions associated with these tiles are treated as deterministic as well.
\end{rem}

For a blind tile $v_i \in V_{\rm blind}$, uncertainty in the local dynamics may cause the system to exit through a facet different from the intended one. The perceived transition probability to a neighboring tile $v_j$ is denoted by $p_{ij}^a = P(v_j|v_i, a)$, where $p_{ij}^a$ is obtained from the facet-wise probability calculation in Section~\ref{subsec:riskUnintended}. The possible successors include the nominal successor $\tau(v_i, a)$ and the neighboring tiles associated with unintended exit facets. 

The traverse cost $c(v_i, a, v_j)$ is defined according to the observation condition. For a deterministic transition from an observed tile, the cost is given by the estimated bound on reach time and possibility of information gain introduced in Section~\ref{subsec:graphConstruction}. For an action originating from a blind tile, let $T_i^a$ denote the estimated bound on reaching time associated with the intended transition. For an unintended successor $v_j$, we define $d_{\rm obs}(v_j) = \min_{v_o \in V_{\rm obs}}d(v_j, v_o)$, where $d(v_j, v_o)$ measures the distance from $v_j$ to an observed tile $v_o$. The transition cost is then written as
\begin{equation}\label{eq:costUnobserved}
c(v_i, a, v_j)=
\begin{cases}
T_i^a, & v_i \in V_{\rm blind}, v_j = \tau(v_i, a),\\
\lambda d_{\rm obs}(v_j), & v_i \in V_{\rm blind}, v_j \neq \tau(v_i, a),
\end{cases}
\end{equation}
where $\lambda>0$ determines the relative penalty assigned to recovery from blindness after an unintended transition. A larger value of $\lambda$ assigns a higher cost to unintended transitions. Consequently, the stochastic transition system accounts for both the traversal time of the intended motion and the difficulty of returning to an observed region when the system exists through an unintended facet.
\begin{figure*}[!t]
  \centering
  \includegraphics[width=\textwidth]{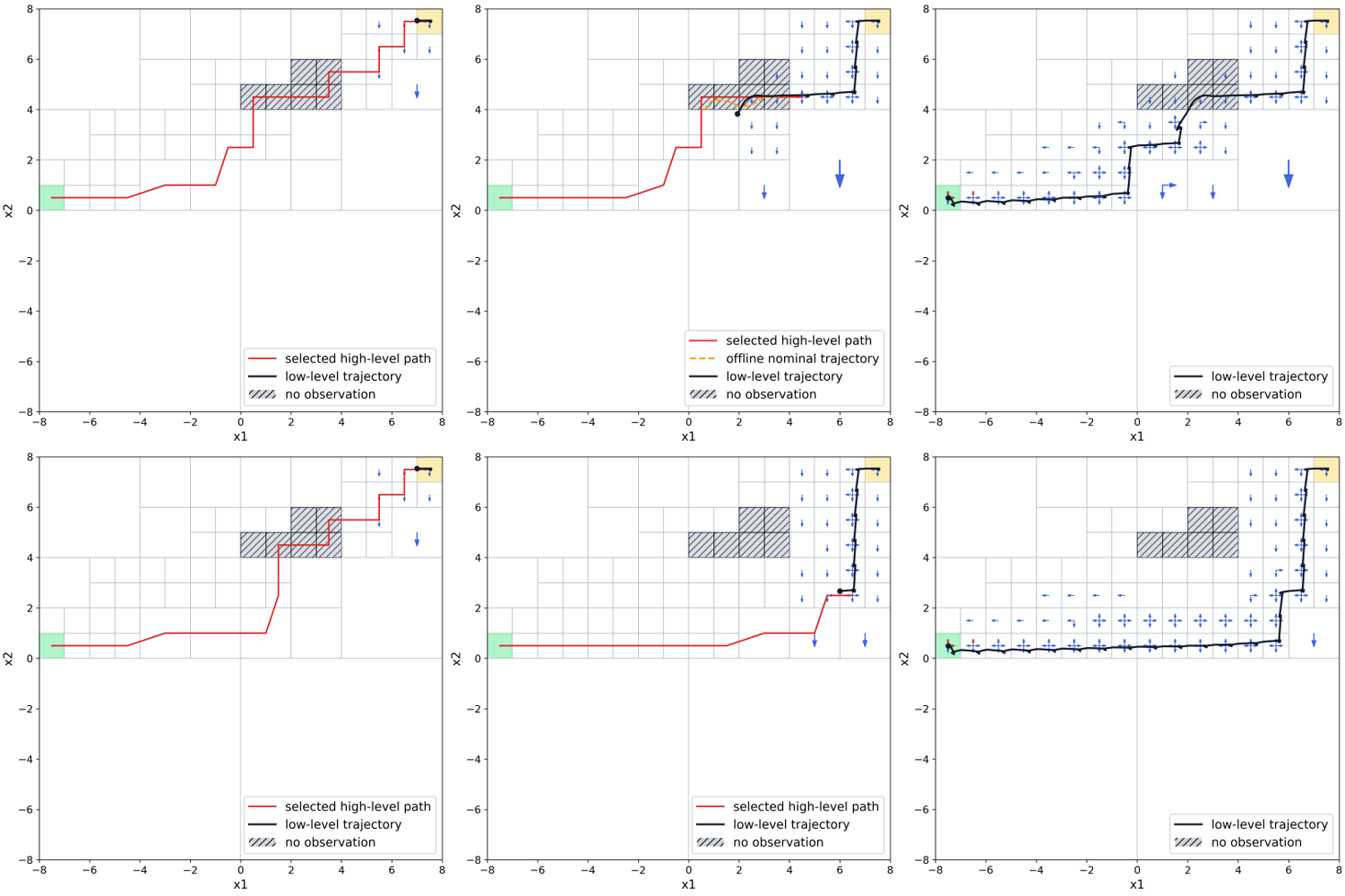} 
  \caption{Illustration of planned policies and corresponding simulated trajectories under two different values of $\lambda$. The yellow tile denotes the tile containing the initial state, and the green tile denotes the tile containing the target state. Shaded tiles represent blind regions. Blue arrows within the tiles represent facet reachability that are certified to exist at the current planning iteration. Red arrows represent facet reachability that are certified to not exist at the same point in time. The red line shows the high-level policy generated by the planner, while the black line shows the actual trajectory. The top row corresponds to the case $\lambda = 1$, and the bottom row corresponds to the case $\lambda = 300$. In each row, the three panels from left to right show the system evolution at three different time instants.}
  \label{fig:1}
\end{figure*}
\subsection{Stochastic Shortest Path Planning}
Based on the stochastic transition system constructed above, the high-level motion planning problem is formulated as an undiscounted stochastic shortest path problem. Let $v_g \in V$ denote the tile containing the target state $x^*$. The target tile $v_g$ is treated as an absorbing terminal state with zero terminal cost.

For each nonterminal tile $v_i \in V\backslash\{v_g\}$, the optimal value function satisfies the Bellman equation
\begin{equation}
    J^*(v_i) = \min_{a\in \mathcal{A}(v_i)} \sum_{v_j \in V}P(v_j|v_i, a)[c(v_i, a, v_j) + J^*(v_j)],
\end{equation}
with $J^*(v_g) = 0$. The corresponding optimal policy is given by
\begin{equation}
    \pi^*(v_i) \in \arg\min_{a\in \mathcal{A}(v_i)} \sum_{v_j \in V} P(v_j|v_i, a)[c(v_i, a, v_j) + J^*(v_j)].
\end{equation}
The Bellman equation can be solved by a standard value iteration procedure. Starting from an initial value function $J_0$, the update is
\begin{equation}
\scalebox{0.8}{$J_{k+1}(v_i) = \min_ {a\in \mathcal{A}(v_i)} \sum_{v_j \in V} P(v_j|v_i, a)[c(v_i, a, v_j) + J_k(v_j)],\ v_i \neq v_j,$}
\end{equation}
while $J_k(v_g) = 0$ is maintained at every iteration. After convergence, the actions that minimize the value function determine the policy $\pi^*$.

The policy is implemented differently depending on the availability of state observations. When the agent is in the observed tile, its action $\pi^*$ can be selected directly and the corresponding reach controller can be synthesized. If the nominal successor of the selected action lies in $V_{\rm blind}$, the nominal transition sequence through the blind region is determined before the state observation is lost. Starting from the current observed tile $v_0^{\rm nom}$, the sequence is generated recursively as
\begin{equation}
    a_k^{\rm nom} = \pi^*(v_r^{\rm nom}),\ v_{k+1}^{nom} = \tau(v_r^{\rm nom}, a_k^{\rm nom}),
\end{equation}
until the first nominal successor $v_K^{\rm nom} \in V_{\rm obs}$ appears in the sequence.

The controllers associated with the resulting nominal sequence are then converted into a single time-indexed open-loop control sequence before the agent enters the blind region. The schedule contains the control inputs for all consecutive blind-tile transitions together with the prescribed switching between the corresponding local controllers. It is executed without state measurements until observation becomes available again. Once the agent reaches an observed tile, the current continuous state is measured, the corresponding tile is identified, and the SSP planning procedure is repeated from the newly observed state.

\section{CASE STUDIES}\label{sec:caseStudy}

In this section, we consider a planar quadrotor navigation problem to illustrate the proposed method.  The vehicle is assumed to have an onboard low-level controller that maintains altitude and tracks planar velocity commands. Let $x = [x_1, x_2]^\top$ denote its planar position and $u=[u_1, u_2]^\top$ denote the velocity command generated by the high-level controller. If the command is transmitted and tracked exactly, the planar motion can be represented by a fully actuated single-integrator model $\dot x = u$~\cite{9131842}. 

We assume that the velocity command generated by the high-level controller is distorted before it is received by the onboard low-level controller. This distortion results in an unknown state-dependent mismatch between the commanded and received velocities. The mismatch is represented by the following nonlinear control-affine dynamics.
\begin{equation}\label{eq:dynamicsDrone}
    \begin{aligned}
        \dot x &= f(x) + g(x)u,\\
        f(x)&= \begin{bmatrix}
            -0.5\sin(0.1x_1 - 0.2 x_2) - 4.5\\
            -0.2\sin(0.3x_1 - 0.1 x_2) - 4.5
        \end{bmatrix},\\
                g(x)&= \begin{bmatrix}
            1+0.02x_1 & 0.02x_2\\
            0.02x_1 & -1+0.02x_2
        \end{bmatrix},\\
    \end{aligned}
\end{equation}
The function $f(x)$ represents additive command biases, while $g(x)$ changes the magnitude or direction of the transmitted commands. When $f(x) = 0$ and $g(x) = \mathbb{I}_2$, the received command equals the transmitted command, and the model reduces to $\dot x = u$.

The functions $f(x)$ and $g(x)$ are unknown to the planner. Only their conservative Lipschitz constants are available with $L_{df} = 0.045$ and $L_g = 0.03$. The workspace is constrained by $-8{\rm m} \leq x_1 \leq 8{\rm m}$, $-8{\rm m} \leq x_2 \leq 8{\rm m}$, and the command velocities satisfy $-5{\rm m/s} \leq u_1 \leq 5{\rm m/s}$, $-5{\rm m/s} \leq u_2 \leq 5{\rm m/s}$. The workspace contains both observed and blind regions. State measurements $x$ are unavailable in blind regions. The state space is partitioned using the non-uniform partitioning procedure described in Section~\ref{subsec:nonuniformPartition}, with the minimum tile size of $1{\rm m}$. The parameters associated with uncertain edges in~\eqref{eq:uncertainWeights} are set to $C_u = 8$ and $\beta_u = 0.4$. The optimization problem in~\eqref{eq:bigOptimization} is nonconvex due to the bilinear terms in its constraints. We use the SLSQP solver implemented in SciPy~\cite{virtanen2020scipy} to seek a locally optimal solution.

To examine the effect of parameter $\lambda$ in~\eqref{eq:costUnobserved}, we compare two values of $\lambda$: $\lambda = 1$ and $\lambda = 300$. A smaller value of $\lambda$ assigns a lower cost to unintended transitions and therefore reduces the expected cost associated with the corresponding action.

As shown in Fig.~\ref{fig:1}, the simulation results show that the proposed method successfully computes a policy from the initial state to the target state and synthesizes the corresponding control inputs. For $\lambda = 1$, the selected route passes through the blind region with a smaller traversal time (about 361 seconds). For $\lambda = 300$, the planner chooses a route that avoids the blind region with a larger traversal time (about 411 seconds). The comparison shows that parameter $\lambda$ balances traversal efficiency against the difficulty of recovering state observation after an unintended transition.

The simulations are implemented in Python using PyBullet~\cite{coumans2021pybullet} on a Windows 11 laptop equipped with an Intel Core i7-13700HX CPU and 16 GB of RAM.

\section{CONCLUSION}\label{sec:conclusion}
In this paper, we consider robotic motion planning and control under unknown dynamics with hybrid state observations. For blind regions, nominal dynamics are selected to synthesize open-loop control sequences, and the possible unintended tile transitions caused by model mismatch between the nominal and true dynamics are quantitatively evaluated. These risks are incorporated into a stochastic transition system, and the high-level planning problem is solved as a stochastic shortest path problem. The case study shows that the proposed method can guide the robot to the target while balancing route efficiency and the risk associated with traversing blind regions.


%%%%%%%%%%%%%%%%%%%%%%%%%%%%%%%%%%%%%%%%%%%%%%%%%%%%%%%%%%%%%%%%%%%%%%%%%%%%%%%%

%%%%%%%%%%%%%%%%%%%%%%%%%%%%%%%%%%%%%%%%%%%%%%%%%%%%%%%%%%%%%%%%%%%%%%%%%%%%%%%%

%%%%%%%%%%%%%%%%%%%%%%%%%%%%%%%%%%%%%%%%%%%%%%%%%%%%%%%%%%%%%%%%%%%%%%%%%%%%%%%%



%%%%%%%%%%%%%%%%%%%%%%%%%%%%%%%%%%%%%%%%%%%%%%%%%%%%%%%%%%%%%%%%%%%%%%%%%%%%%%%%

\bibliographystyle{IEEEtran}
\bibliography{ref}

\end{document}